\documentclass[12pt]{article}
\usepackage{graphicx}
\usepackage{times}
\usepackage{pdfpages}
\newenvironment{sciabstract}{%
\begin{quote} \bf}
{\end{quote}}

\newcounter{lastnote}

\title{Mapping the world population one building at a time}

\author
{Tobias G. Tiecke$^{1 \dag}$, Xianming Liu$^{1}$, Amy Zhang$^{1}$, Andreas Gros$^{1}$, \\
Nan Li$^{1}$, Gregory Yetman$^{2}$, Talip Kilic$^{3}$, Siobhan Murray$^{4}$, \\
Brian Blankespoor$^{4}$, Espen B. Prydz$^{3}$, Hai-Anh H. Dang$^{4}$\\\\
\normalsize{$^{1}$Facebook Inc., 1 Hacker Way, Menlo Park, CA 94025, USA}\\
\normalsize{$^{2}$Center for International Earth Science Information Network (CIESIN)}\\
\normalsize{Columbia University, Palisades, NY 10964, USA}\\
\normalsize{$^{3}$Development Data Group, The World Bank, Rome, Italy}\\
\normalsize{$^{4}$Development Data Group, The World Bank, Washington DC, USA}\\
\normalsize{$^\dag$To whom correspondence should be addressed. Email: ttiecke@fb.com}
}

\date{}

\begin{document}

\baselineskip24pt

\maketitle
\begin{sciabstract}
High resolution datasets of population density which accurately map sparsely-distributed human populations do not exist at a global scale. Typically, population data is obtained using censuses and statistical modeling. More recently, methods using remotely-sensed data have emerged, capable of effectively identifying urbanized areas. Obtaining high accuracy in estimation of population distribution in rural areas remains a very challenging task due to the simultaneous requirements of sufficient sensitivity and resolution to detect very sparse populations through remote sensing as well as reliable performance at a global scale. Here, we present a computer vision method based on machine learning to create population maps from satellite imagery at a global scale, with a spatial sensitivity corresponding to individual buildings and suitable for global deployment. By combining this settlement data with census data, we create population maps with $\sim 30$ meter resolution for $18$ countries. We validate our method, and find that the building identification has an average precision and recall of $0.95$ and $0.91$, respectively and that the population estimates have a standard error of a factor $\sim 2$ or less. Based on our data, we analyze $29$ percent of the world population, and show that $99$ percent lives within $36$ km of the nearest urban cluster. The resulting high-resolution population datasets have applications in infrastructure planning, vaccination campaign planning, disaster response efforts and risk analysis such as high accuracy flood risk analysis.
\end{sciabstract}

Accurate information on global population distribution is crucial to many disciplines. A population and housing census is the traditional tool for deriving small-area detailed statistics on population and its spatial distribution\cite{UN}. However, censuses are time-consuming, and the spatial resolution is naturally set by the census enumeration areas (EA), which lack fine-grained information about the aggregation of population.
The sizes of the EAs vary by many orders of magnitude from country to country, ranging from hundreds of square meters in urban areas to tens of thousands of square kilometers in low population areas, resulting in an average spatial resolution\cite{balk06} of a census unit of 33 km at a global scale. 
Recently, multiple higher resolution maps of human-made built up areas have emerged~\cite{potere09,gamba09}, most notably the Global Human Settlement Layer (GHSL)\cite{pesaresi15,pesaresi16}, the Global Urban Footprint (GUF)\cite{guf1,guf2}, the WorldPop project\cite{worldpop}, Landscan\cite{landscanhd} and Missing Maps project\cite{missingmaps}. However, none provide a scalable solution with high accuracy in rural areas.

Over the past decade high-resolution (sub-meter) satellite imagery has become widely available, enabling the global collection of recent and cloud-free earth imagery. Additionally, in the past years, the surge in research on computer vision and in particular convolutional neural networks (CNN) have enabled bulk processing of imagery in a rapid manner \cite{krizhevsky2012imagenet,szegedy2015going,he2016deep}. The combination of these methods enables the global analysis of high-resolution imagery as a promising method 
for detecting individual buildings; combining building estimates with available census data to produce updated and higher resolution population maps; and offering alternative, state-of-the-art population estimates in the absence of census data. Various approaches using machine learning have been demonstrated on small areas~\cite{pacifici2009,yuan2016automatic}, yet a method which allows global mapping has remained elusive.

Here, we present a method to generate high-resolution population density maps at a global scale (see figure 1) (referred to as \emph{High Resolution Settlement Layer (HRSL)}). By using various CNN architectures, we identify human-made buildings in high-resolution satellite imagery, for a wide range of terrains, seasons and climates, even under poor image-quality conditions. Under the assumption that buildings act as a proxy for where people live, we obtain population estimates on a country-wide level, with 1x1 arcsecond resolution ($\sim 30\times30\,\mathrm{m}$ at the equator) and sensitivity to individual buildings, enabling accurate studies of population aggregation in rural areas. 

To enable global analysis we develop a building detection model that is country-agnostic, fast, and works with easily obtainable binary labeled data. Our pipeline consists of several steps; we extract 64x64 pixel images ('\emph{patches}') around detected straight lines using a conventional edge detector, which reduces the amount of data for classification by $\sim 4$ times.
A portion of those candidates are sampled across all countries and labeled as training and evaluation data for the CNNs (see Supplemental Information (SI)). The computer vision models are trained on a single machine with four GPUs, whereas the classification runs on Facebook's infrastructure on a CPU cluster. This approach allows us to process $0.5\,\mathrm{M km^2}$ in $\sim 24$ hours. 

We use three different types of CNN (see figure 2): a classification model based on the SegNet~\cite{segnet} architecture in order to perform binary classification (building/no-building); a feedback neural network (FeedbackNet) performing weakly-supervised segmentation of the satellite images enabling us to obtain building footprints~\cite{feedbacknet}, and a denoising network which is capable of improving the quality of the source data by removing high-frequency noise from the satellite imagery (see SI). For all models the tradeoff between good performance at a global scale (larger networks) and the run time of the networks to perform global analysis (smaller networks) determined the limit in performance of the models. 

The encoder-decoder style SegNet, is customized to perform the classification at the level of a \emph{patch}. The encoder (a convolutional sub-network) is used to extract abstract information about the input, and the decoder (a deconvolutional sub-network) is trained to upsample the output of the encoder into a spatially meaningful probability map representing the possibility of house existence in the input. The probabilities generated by the decoder are averaged over all spatial locations within the patch to derive the final classification. This yields high accuracy and a reduced false positive rate on a global scale compared to other methods we explored, however, at the expense of loss of spatial details such as the boundaries and shapes of individual buildings.

To recover the building shapes in addition to classification we use a separate CNN to perform image semantic segmentation~\cite{long2015fully, chen2014semantic}.
Traditional methods, such as region proposal~\cite{uijlings2013selective} based CNNs~\cite{ren2015faster}, are not able to handle small-sized foreground objects. Moreover, these methods require a large amount of pixel-wise labeled training samples following a supervised learning setting.
Obtaining such a large volume of labeled data is currently intractable to achieve at a global scale because of it being time-consuming to acquire, as well as it having a large possibility to accumulate errors in supervision. To facilitate a generalized and scalable model, we employ the weakly-supervised learning that takes the abundant and easy-to-get image level categorical supervision (binary labeling) into training, and perform pixel-level prediction during deployment~\cite{feedbacknet}. 
The methodology is motivated by the feedback mechanism in human cognition~\cite{desimone1995neural} and recent advances of computational models in Feedback Neural Networks~\cite{cao2015look}, which deactivates the non-relevant neurons within hidden layers of neural networks and achieve pixel-wise semantic segmentation.
Both models are trained on $150,000$ binary labeled (building/no building) \emph{patches}, randomly sampled from all countries and seasons, covering both rural areas and urban areas. 

We have ran our model on 18 countries across 3 continents, spanning 29\% of the world's population. For nearly all countries we analyze over 95\% of the landmass, of which $1.9\%$ contains a building at a $1\times 1$ arcsecond resolution. The coverage is limited by the availability of cloud-free, high-resolution satellite imagery (see SI) and nosignificant differences in performance between countries is observed, unless the source imagery is of poor quality.

A well characterized accuracy of the dataset is crucial for its further use. We consider the accuracy of the building identification and the accuracy of the population redistribution over buildings separately, with a focus on the former, which is the main topic of our work; the latter will be addressed briefly.
Since a global ground truth dataset is not available four independent analyses are performed, each testing different potential errors in our dataset.
Statistical errors and systematic errors are treated separately, the former having random origin and the latter originate e.g. from repetitive errors, such as large rocks, boats or mountain ridges (see Figure 3) being falsely classified as buildings. Systematic errors are more challenging to quantify but are of particular concern since they can result in clusters of false positives potentially misinterpreted as a settlement. 
To quantify statistical errors, we first study the accuracy of the computer vision model by comparing our classification results with a reference (human labeled) dataset. Averaged over 18 countries the SegNet achieves a precision and recall of $Pr=(TP/(TP+FP)) = 0.947$, $Re=(TP/(TP+FN))=0.913$ ($TP, FP, FN$ are true positive, false positive and false negative counts respectively) on a global dataset where only $7\%$ of the randomly sampled patches contains a building. This analysis tests the performance of our model, however, it doesn't capture errors due to clouds, missed straight edge detection or missing source imagery. 

To address this, a second analysis is performed comparing the identified buildings to the GPS coordinates of the households interviewed for the Malawi Third Integrated Household Survey (IHS3). This dataset is obtained by interviewing a nationally-representative sample of households, and thus, is independent of artifacts of remote sensing.
The incidence of finding a household is measured within a distance of at most 100m from the nearest populated pixel in our building dataset, thereby accounting for the limited accuracy of the survey location data (see SI). At the national-level, $98.3$ percent of the $11,957$ IHS3 households coincide with a populated pixel in our building dataset. 

In order to quantify the systematic errors we compare our data against two datasets (GUF and GHSL) generated using independent data sources and methodologies. The three datasets are compared at the 1 arc-second resolution of the binary classification to create all possible combinations of absence or presence of a settled area (see figure 3c).
The areas where our classification disagrees with both the GUF and GHSL classifications represented as contiguous vectors are visually inspected using the satellite imagery to identify false positive and negative units. Country totals of agreement and disagreement as a proportion of country area are shown in Table 1 (see SI for additional countries). Four test cases of Malawi, Ghana, Vietnam and Haiti are presented since they cover a wide range of geographies and image quality.

We find that the visually confirmed systematic errors contribute to less than 1\% of the total landmass and only a few percent of the country built up area. Additionally, in general areas of systematic false negatives occur more often than areas of systematic false positives. In order to assess the impact of larger clusters of disagreement which could yield more severe systematic areas we consider the $\sim 500$ largest areas of disagreement. These errors contribute to less than $1\%$ of the built up area for all 4 countries. The GUF and GHSL often miss buildings which are discovered in ours. 

Finally, we perform an analysis comparing the three datasets (HRSL, GUF, GHSL) against a single region near Blantyre, Malawi which has been completely manually labeled, at a resolution of 2x2 arcsecond to allow for misalignment between the datasets (see figure 3c)\cite{missingmaps}. The northern urban area (above the upper dotted line) is captured fairly well in all three datasets expressed in the recall values of $0.83/0.82/0.99$ for the GHSL/GUF and HRSL respectively. However, for the southern rural area (below the lower dotted line) the recall values are $0.04/0.06/0.84$ for the GHSL/GUF and HRSL respectively, demonstrating the superior accuracy of our dataset in rural areas. 
In conclusion, all errors are estimated to be a few percent or less, indicating a high degree of confidence in the building dataset. Additionally, our results are superior over existing datasets in rural areas.

After assessing the accuracy of the buildings, we turn to the population redistribution over the buildings. Population estimates are obtained using a minimally modeled approach of proportional allocation \cite{ciesinpopalloc,doxsey}. We take two approaches, first, the population is distributed equally to settled areas in the binary classification that fall inside census units and second, the population is distributed proportionally to the fraction of built up area within a 1 arcsecond gridcell (see SI for details)\cite{policy}. The population allocations of the second method are expected to be more accurate, however, since currently no method exists to validate the population data globally at sub-arcsecond lengthscales we limit our analysis to the first method.
The largest uncertainty of the population estimates originates from census data obtained at a too coarse administrative unit. The impact is estimated by comparing the population estimates performed on a coarse administrative unit to known population counts of finer administrative units (see SI). The error (standard deviation) in population estimate is a factor of 2 or less depending on the available country specific census data. Finally, the errors are slightly larger in urban areas (2.5 or less) compared to rural areas. Our data is well suited for a more sophisticated population allocation, which is an active area of research\cite{worldpop}, however, this is beyond the scope of this work.

Our results represent a large improvement over the state-of-the art in globally robust building identification via satellite imagery, and thus population estimation, particularly for rural areas. As mentioned in the introduction, population estimates are a crucial interdisciplinary concern, and a cursory evaluation of our data hints at the nature of insights to come. For example, the improved rural accuracy enables the study the population distribution with respect to the nearest urbanized center. Definitions of urban versus rural vary widely, and are an active research area. Using the definition of Ref. \cite{ghsurban} for \emph{urban clusters}; clusters with a population density of at least $300\,\mathrm{/km^2}$ and a minimum population of $5000$ we calculate the distribution of the rural population as a function of the distance to the nearest urban gridcell. Figure 4 shows the distribution of the $18$ countries we have analyzed, and shows that $90\% / 95\% / 99\%$ of this population lives within $7/14/36\,\mathrm{km}$ of the nearest urban cluster. This proximity of rural population to urban clusters provides guidelines for telecommunication infrastructure development.

In conclusion, our model achieves high accuracy and is applicable at a global scale. The errors in building identification are estimated to be at the few percent level on a nation wide scale, while maintaining an unprecedented sensitivity to individual buildings. The uncertainty in population estimates originates from the underlying census data and the method of proportional allocation which is a topic of future research.

\begin{center}
\begin{figure}[htbp]
\includegraphics[width=7in]{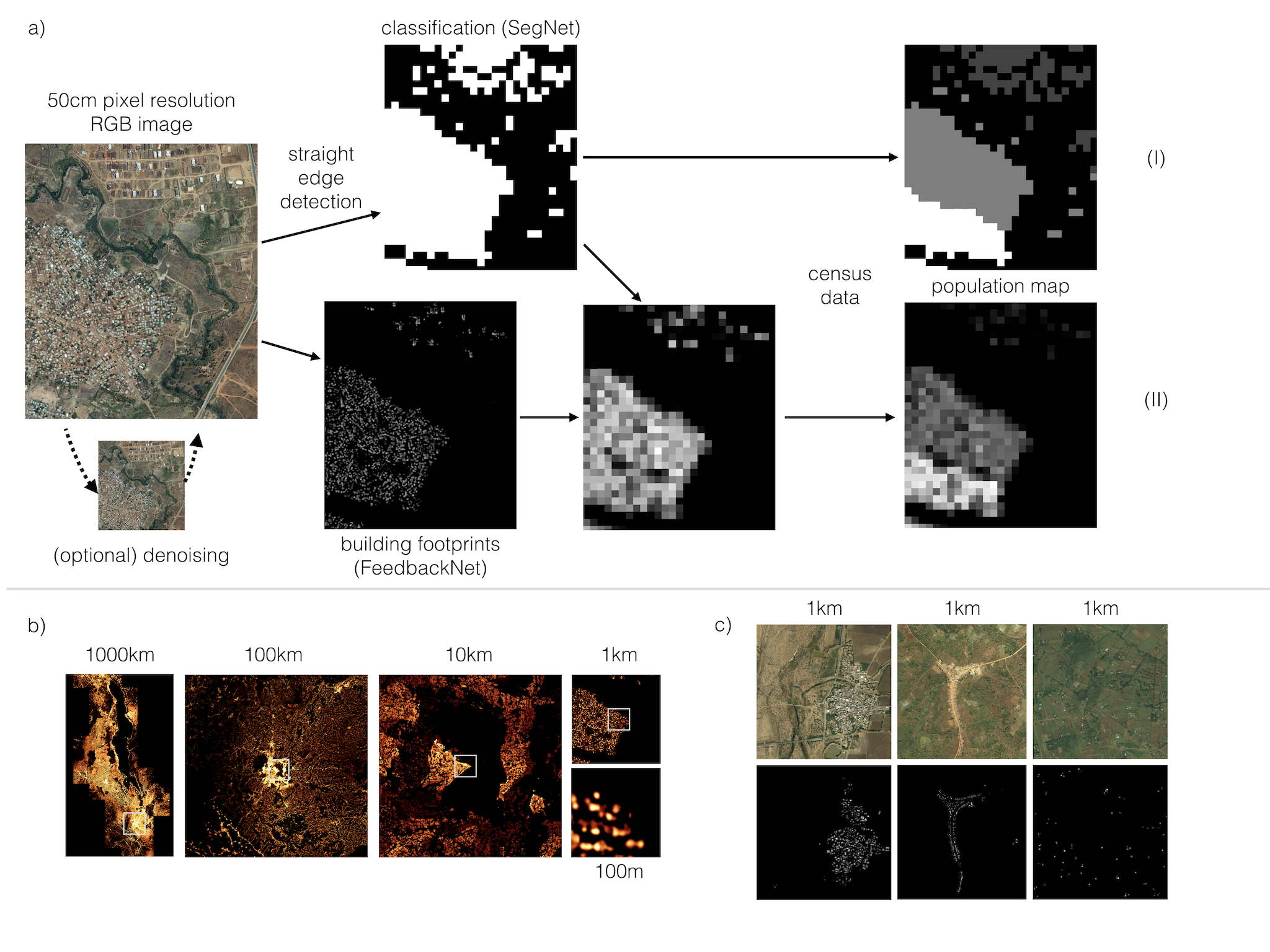}
\caption{\label{fig:fig1}
  \textbf{Method to create population maps from high resolution satellite imagery.} \textbf{(a)} Our method starts with satellite imagery to which a denoising network is applied depending on the image quality. Subsequently, the image is analyzed using straight edge detection and two independent CNNs, SegNet and FeedbackNet, resulting in accurate classification and determination of the building footprint. Two independent approachesare studied to create a built up area map (I) uses classification results and for (II) cascades the CNN outputs where each gridcell corresponds to the fraction of built up area. By combining this with census data a population estimate at 1x1 arcsecond resolution.  \textbf{(b)} Examples of the results for Malawi covering 5 orders of magnitude in length scales. \textbf{(c)} Examples of the model output for (left-right) India, Mozambique and Kenya.}
\end{figure}
\end{center}

\begin{center}
\begin{figure}[htbp]
\includegraphics[width=7in]{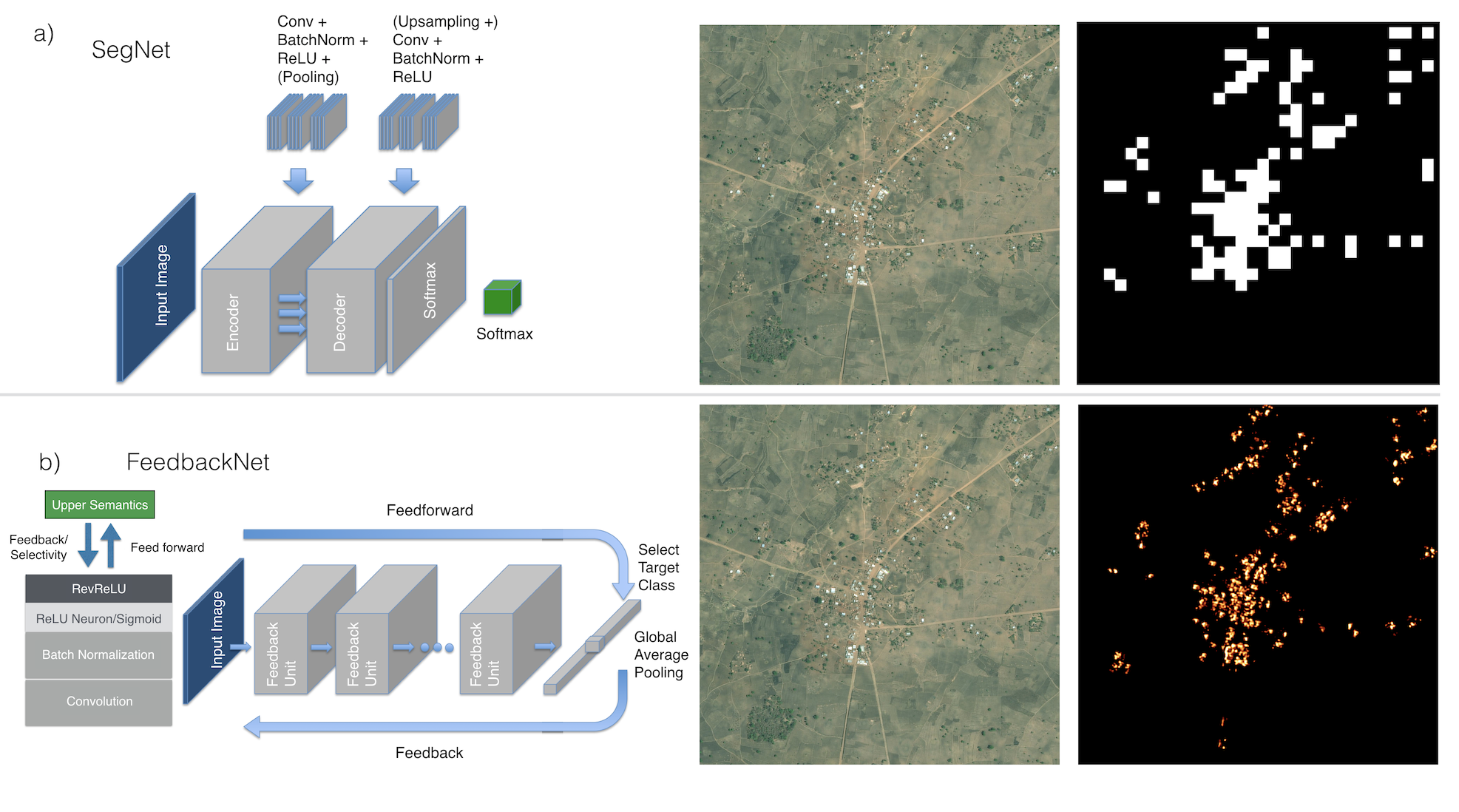}
\caption{\label{fig:fig2}
\textbf{CNNs for building detection} \textbf{(a)} for classification of \emph{patches} an architecture based on SegNet\cite{segnet} is used. \textbf{(b)} for image segmentation of the built up area within a \emph{patch} we use a weakly-supervised FeedbackNet\cite{feedbacknet} architecture on 256x256 pixel images. This network outputs a feature map indicating the building footprint. Since this network is pixel-based it has more false positives than SegNet, which is accounted for by cascading the outputs of the two models.}
\end{figure}
\end{center}

\begin{center}
\begin{figure}[htbp]
\includegraphics[width=7in]{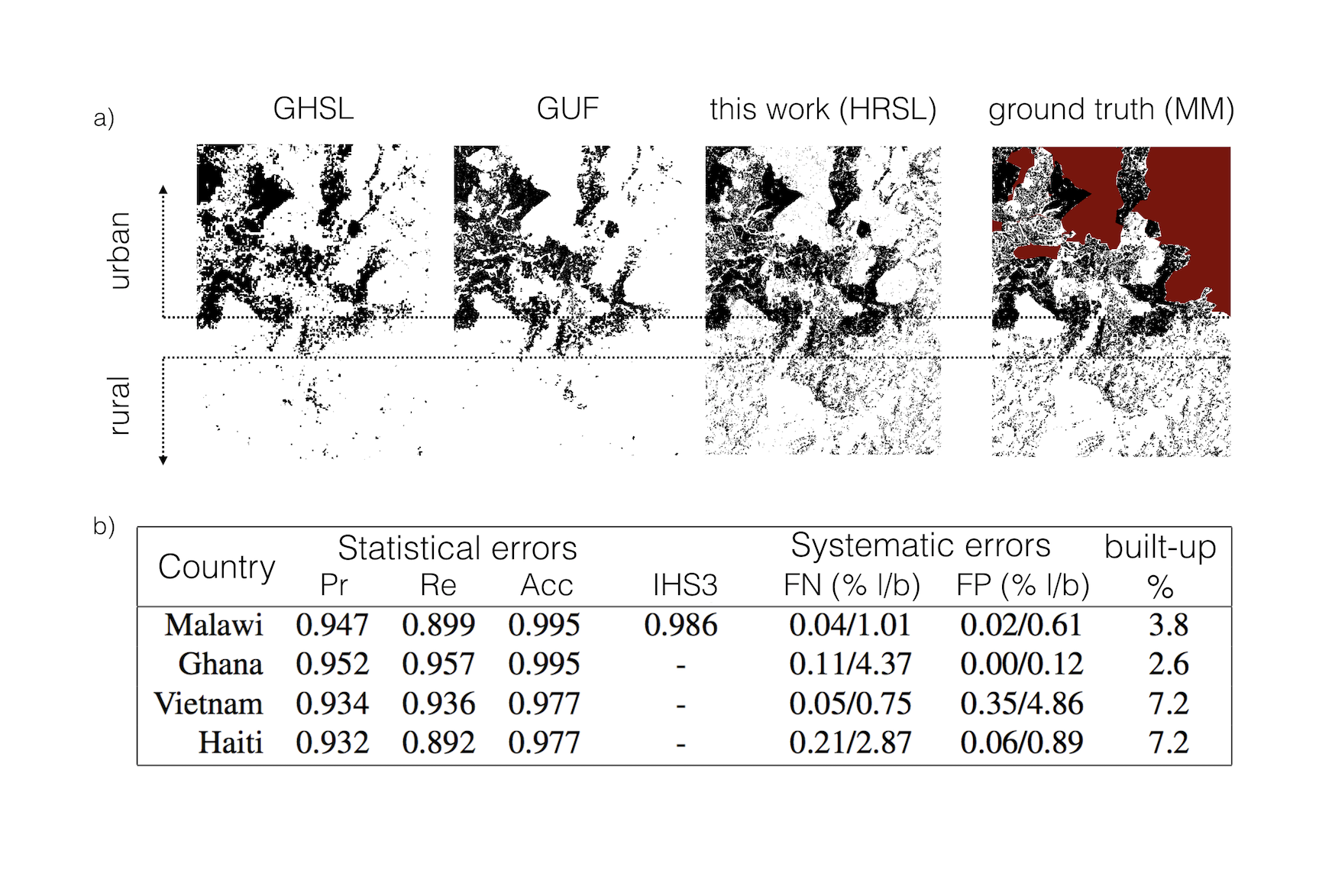}
\caption{\label{fig:fig3}
\textbf{Validation and error analysis} \textbf{(a)} comparison of GUF, GHSL, HRSL and human labeled ground truth (Missing Maps project, not mapped areas are indicated red). Our dataset accurately captures rural areas in the southern part of this region which are omitted in the GUF and GHSL methods (see text). \textbf{(b)} statistical and systematic errors, the false negative (FN) and false positive (FP) values are shown as percentage of the total landmass (l) and total built up area (b).}
\end{figure}
\end{center}

\begin{center}
\begin{figure}[htbp]
\includegraphics[width=6in]{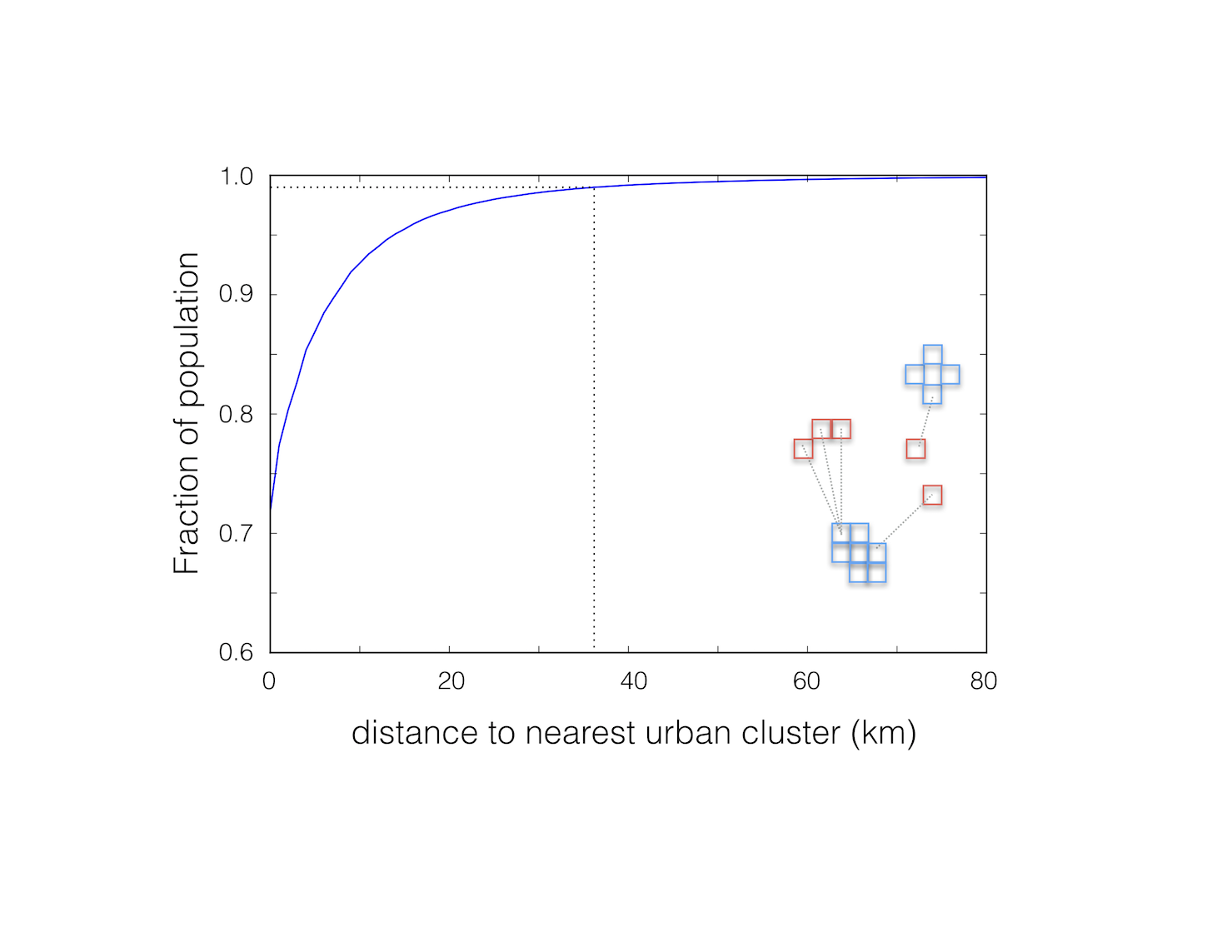}
\caption{\label{fig:fig4}
\textbf{Distribution of population near urban clusters.}  By clustering the population estimates urban clusters are identified and for all population the distance to the nearest urban cluster is determined. Of the studied population ($18$ countries, covering $29\%$ of the world population) $99\%$ of lives in or within $36\,\mathrm{km}$ of an urban cluster. (inset) shows a schematic of the distance calculation; 1x1km gridcells are classified as an urban cluster (blue) or rural (red). The distance is calculated for all population to the nearest urbanized gridcell.}
\end{figure}
\end{center}

\section*{Acknowledgement}
We thank D. Liu and T. Huang for insightful discussions on the denoising and feedback networks, and C. Deuskar and B. Stewart for insightful discussions on urbanization. All satellite imagery \textcopyright  2016 DigitalGlobe. Funding for World Bank affiliates has been provided by The World Bank Living Standards Measurement Study ? Integrated Surveys on Agriculture (LSMS-ISA) project, and
The World Bank Strategic Research Program Projects: (i) Use of Satellite Data and CDRs for Census-Independent Spatial Distribution of Populations and (ii) Survey Imputation for Improved Poverty and Shared Prosperity Diagnostics. Funding for CIESIN affiliates has been provided by Facebook. The boundary and census data collection were developed with funding from the National Aeronautics and Space Administration under Contract NNG13HQ04C for the Socioeconomic Data and Applications Distributed Active Archive Center (DAAC). Funding for Facebook affiliates was provided by Facebook, Inc.



\section*{Supplementary material}

# Supplementary Information for
*Mapping the world population one building at a time*

1. **Source satellite imagery**

    1.1. **Availability and requirements**

Our models are based on satellite imagery from Digital Globe. The data is largely from the Vivid+ product, augmented with grayscale data in areas with high cloud cover and in very rare cases the Global Basemap RGB data. All data has 50cm resolution, which is required to achieve sensitivity to individual houses. We tested the sensitivity of our approach on the resolution of the imagery by repeating our approach on downsampled imagery, which is subsequently re-usampled by applying the super resolution technique [1] to reconstruct the high resolution imagery. We find that individual houses can still be recognized when downsampling by 3x (1.5m resolution), albeit with reduced accuracy. At 2m resolution and less individual houses are not recognized anymore. At the time of writing, the only commercially available product with spatial resolution of 1.5m or better and (near) global coverage is provided by Digital Globe.

In order to achieve cloud free coverage, the satellite imagery naturally is composed of images taken over a spread of time, typically multiple years per country. Of the data used for the countries listed below, 90% was taken between 2011 and 2015. Further limitations in the performance arise from seasonal differences, however, when sufficient data of different seasons is included in the training dataset this is mitigated.

   1.2. **Denoising neural network**

Noise in source images is a big concern for a convolutional neural network to generate accurate predictions[2]. Unfortunately, in a mosaic of satellite images a certain fraction of the images are noisy due to the normalization or calibration in pre-processing steps, this leads to a spatial dependency of the noise with can lead to spatially determined systematic errors. To handle this problem considering both efficiency and effectiveness, we implemented an Image Denoising Neural Network (DeepDenoiser), as shown in Figure S1a.

The network is trained end-to-end in a regression setting, with a generator producing random noise applied to original input dynamically. Compared with the state-of-the-art (BM3D[3]), our approach does not require a large search space and can achieve similar performance with at least 10 times efficiency gains in speed. This allows us to run the denoising at large scale. As shown in Figure S1b, the DeepDenoier improves the data quality significantly compared with the original raw input data; while Figure S1c shows the detection results before and after denoising on Lagos, Nigeria, where the denoising helps find the missed city caused by low data quality.

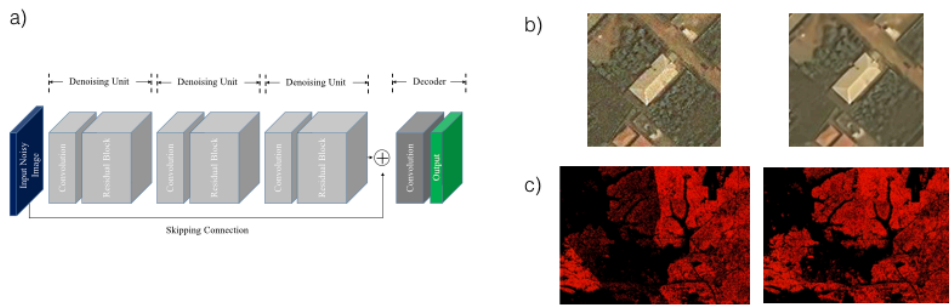

*Figure S1. Image denoiser neural network. a) the network architecture. b) and c) show the performance of the denoiser in Lagos, Nigeria. On the left noisy images result in omission of a large area in the west of Lagos, while after applying the denoising network (right) the missed buildings are recovered. Satellite images © 2016 Digital Globe.*

## 2. Evaluation of neural networks

### 2.1. Training and testing data labeling

To train and evaluate machine learning models for building detection, we sample 150K 64x64 patches detected from straight-line detector to label. For the training data, we adopt a boostrapping method to make positives and negatives balanced. While for testing data, we use randomly sampling over spatial location to ensure the testing data is of the identical distribution with the real-world data.

All samples are labeled by at least two different labelers, and the ones with dispute are removed to ensure the data correctness. Even though, we still observe large portion of incorrectness, especially in rural areas, where the imagery data quality is usually bad and and density of buildings is low. As a consequence, labelers tend to label "negative" since there are far more negative examples than positives. Accuracy of single human labeler is calculated by using golden dataset, which is around 85% over all labelers. Considering the accuracy of machine learning model at a global scale being over 90%, we can make a conclusion that the machine learning model is superior than human in performing this particular task.

### 2.2. Precision and recall results by country

In Table S1 we list the precision and recall numbers by country, as well as the fraction of the landmass which is covered in our analysis. Coverage less than 100% is due the lack of the availability of cloud-free satellite imagery.

| Country | Precision | Recall | Coverage (%) | Urban population (%) | p90 (km) | p95 (km) | p99 (km) |
|---|---|---|---|---|---|---|---|
| Algeria | 0.948 | 0.913 | 100 | 66.5 | 10.3 | 17.9 | 67.7 |
| Burkina Faso | 0.940 | 0.919 | 100 | 33.2 | 26.6 | 33.2 | 48.2 |
| Cambodia | 0.918 | 0.941 | 97.2 | 51.1 | 16.4 | 28.2 | 48.8 |
| Ghana | 0.952 | 0.957 | 99.0 | 56.8 | 40.6 | 127.5 | 171.9 |
| Haiti | 0.932 | 0.892 | 100 | 11.6 | 93.7 | 111.5 | 134.0 |
| India | 0.993 | 0.961 | 99.1 | 74.3 | 4.6 | 8.7 | 20.5 |
| Indonesia | 0.944 | 0.883 | 96.8 | 78.3 | 8.3 | 21.3 | 79.5 |
| Ivory Coast | 0.950 | 0.960 | 96.1 | 42.4 | 53.1 | 104.0 | 172.2 |
| Madagascar | 0.926 | 0.862 | 100 | 30.0 | 52.6 | 67.5 | 102.3 |
| Malawi | 0.947 | 0.899 | 100 | 46.3 | 15.8 | 22.4 | 42.4 |
| Nigeria | 0.922 | 0.871 | 97.5 | 63.7 | 16.7 | 32.4 | 80.9 |
| Philippines | 0.954 | 0.901 | 100 | 73.7 | 7.5 | 23.2 | 84.9 |
| Rwanda | 0.912 | 0.912 | 100 | 85.7 | <1.0 | <1.0 | 4.8 |
| South Africa | 0.984 | 0.876 | 100 | 71.3 | 22.5 | 38.4 | 88.9 |
| Sri Lanka | 0.977 | 0.933 | 100 | 73.7 | 8.1 | 41.1 | 72.9 |
| Thailand | 0.931 | 0.892 | 99.2 | 48.4 | 19.9 | 27.4 | 48.2 |
| Uganda | 0.989 | 0.926 | 95.0 | 58.3 | 10.4 | 16.7 | 29.6 |
| Vietnam | 0.934 | 0.936 | 83.0 | 80.5 | 2.5 | 7.2 | 22.3 |

*Table S1. Precision, recall and coverage by country.*

## 3. Allocating population to buildings

Census unit boundaries and associated population estimates from the Gridded Population of the World collection [4], version 4 (GPWv4) are used to distribute population to the binary classification data at a resolution of 1 arc-second, approximately 30x30m at the equator. The census unit boundaries collected for GPWv4 vary in the level of spatial detail and date of estimate; Table S2 shows the administrative level, number of units, average unit size, and census year and for each country. The GPWv4 census data include population estimates for five time periods: 2005, 2010, 2015, and 2010; these estimates are interpolated or extrapolated from at least two census dates for each country[5]. The 2015 estimate is used in the population model to create comparable results across countries.

| Country | Administrative Level | Number of Units | Census Year |
|---|---|---|---|
| Ghana | 2-Parliamentary Constituency | 170 | 2010 |
| Haiti | 4-Section Communale | 560 | 2003 |
| Malawi | 3-Enumeration Area | 12,645 | 2008 |
| Vietnam | 3-District | 688 | 2009 |

*Table S2 Characteristics of source census data. Data sources are as follows: Ghana: [10,11], Haiti: [12,13]; Malawi: [14]; Vietnam: [15].*

Population data are modeled using a minimally modeled approach of proportional allocation [5,6], distributing population equally to settled areas in the binary classification that fall inside census units.[1] We use the minimally modeled approach because it is easily understood, suitable for use as an independent variable in studies that use other physical data, and amenable to additional modeling with additional data. The model is used to create a surface of population counts for each country that sums to match the country population estimate in 2015. The population surfaces represent settled areas with homogeneous population counts within census units. Variability in population distribution are a direct result of settled areas identified in the binary classification and the size of the administrative units. South Africa, with much smaller census units on average, provides a more detailed distribution of population than Ghana.

### 4. Validation of population redistribution

In gauging the accuracy of the population maps generated by the proportional allocation over the building data as described in Section 3, we first assess the maps at the level of the census enumeration area (EA)[2]. The analysis is conducted for Ghana, Malawi, and Vietnam. First, using the EA shapefiles obtained from the respective national statistical agencies, we sum pixel-specific population counts from the high-resolution population maps at the EA-level. Subsequently, we document the census EA-level correspondence between the population estimates according to the population maps and the latest census-based population counts that are updated with country- and location-specific annual population growth rates that are used by CIESIN for generating the GPWv4.[3,4]

Our analysis is based on Ordinary Least Squares (OLS) regressions of the following form:

(1) $y_i = \alpha + \beta P_i + \varepsilon_i$

where *i* denotes the census EA; *y* denotes the 2015 census-based population projection; *P* denotes the population estimate summed across the high-resolution population map pixels at the EA-level, and *α* and *ε* denote constant and error terms, respectively. Equation 1 is estimated in the national,

---

[1] A coastal buffer of 100m is used to include settled land areas that fall slightly outside of the coastline represented in the census units.
[2] Each country is divided into enumeration areas by the respective national statistical office, specifically for the purpose of the population and housing census. An EA could encompass one part of a ward/village, a complete ward/village or a collection of wards/villages, and constitutes the highest-resolution at which aggregate individual and household population numbers are produced from a population and housing census (PHC).
[3] The latest PHC was conducted in 2008, 2009 and 2010 in Malawi, Vietnam, and Ghana, respectively.
[4] The administrative unit level at which CIESIN annual population growth rates are available differs by country. The rates are defined at the traditional authority-level (administrative unit level 2 out of 3) in Malawi; the parliamentary constituency-level (administrative unit level 2 out of 4) in Ghana; and the district-level (administrative unit level 3 out of 4) in Vietnam.

rural, and urban domains, separately, in Malawi and Vietnam[5], and only in the urban domain in Ghana because of the urban-only availability EA-level population counts from the latest census.[6]

Given the cross-country differences in the resolution of the census-based population input data underlying the high-resolution population maps, the selected countries provide an opportunity to explore the implications of such variation on the EA-level correspondence between the census-based population projection and the Facebook population estimate.

In the specific case of Malawi, the census-based population input data for the high-resolution map is already provided by the Malawi National Statistical Office (NSO) to CIESIN at the EA-level, i.e., the highest resolution at which input data could be defined. For this reason, we generate two simulated versions of the population estimates in Malawi – under the artificially-created scenarios of using population input data defined at the traditional authority (TA)-level (administrative unit level 2) and, separately, at the district-level (administrative unit level 1), as opposed to at the EA-level (administrative unit level 3).[7]

The results of the regression analyses are summarized in Table S3. Each estimation of Equation 1 uses the population estimate, and the resolution of the available census population data underlying the high-resolution population map. We report the $R^2$ value associated with each estimation. We have performed this analysis for allocating the population proportional to the binary classification within 1x1 arcsecond gridcell as well as proportional to the builtup fraction within that gridcell (see main text). We find no significant differences between both methods at the lengthscales of administrative unit as performed in this analysis and therefore restrict ourselves to the binary classification results.

Under the simulated population input data scenarios in Malawi, we see that population input data defined at the traditional authority-level or district-level would have a significant bearing on the accuracy of the population estimate. For example, across the nation as a whole and under the TA-level population input data scenario, the EA-level population estimate would explain 37.5 percent of the variation in the census-based population projection. The comparable statistic under the even coarser district-level population input data scenario would decline to 21.8 percent.

These results provide insight in the errors induced by the of the coarseness of the source population data used to allocate to our buildings data. At the national-level, the reduction of the $R^2$ values for Malawi correspond to an estimated error in the population data (standard deviation) of a factor $\sigma_{TA}$

---

[5] In Vietnam, the analysis is restricted to the EAs whose center point latitudes were greater than 10 and less than 17.3. This restriction leads us to work in central Vietnam; in an area in which the built-up area estimates were derived with confidence. The satellite imagery for the rest of the country was otherwise not useable due to perpetual cloud cover. A negligible number of observations were dropped due to the lack of correspondence between the shapefiles and the census location information.

[6] The urban/rural attribute is defined at the EA-level, based on the population and housing census.

[7] To derive the first version, we compute the total percent built-up area in each traditional authority; divide the cell-level percent built-up area by the total traditional authority-level built up area to compute the cell-level allocation ratio; compute the total population estimate in each traditional authority; multiply each cell's allocation ratio by the total traditional authority-level population; and compute the new EA-level population estimate by summing across the new population estimates across the cells in each EA. The second version is computed in an identical fashion, with traditional authority-related parts of the computation above replaced with the computations at the district-level.

= 1/sqrt(0.415) = 1.6 and $\sigma_D$ = 1/sqrt(0.232) = 2.1 for the TA and district-level population input data respectively. We also observe a slightly worse performance in urban areas compared to rural areas. We attribute this to the expected larger diversity of building heights and usage (residential versus non-residential) in urban areas. We illustrate this difference further by looking at the errors in population allocation as a function of EA size. Figure S2 focuses on Accra Metropolitan Area in Ghana. Panel A depicts the extent of EA-level difference between our population data and independent census-based population projection. We see a clear pattern of over–allocation of population to larger (presumably less dense) EAs, and under-allocation to smaller EAs, likely because our model does not account for building height and therefore it does not properly account for changes in population per structure.

Overall, this analysis shows that the accuracy population estimates is limited by the available population input data and its redistribution rather than our building identification. Future research will focus on how to more accurately distribute population counts to building structures by, for example, classification of the building type as well as third-party geospatial data, including those on agroecology, climate, and land use.

| Domain | Country | Level of Administrative Unit for Input Data † | Level of Administrative Unit for Validation Data | Observations | $R^2$ |
|---|---|---|---|---|---|
| National | Vietnam | 3 | 4 | 3,170 | 0.508 |
|  | Malawi | 2 | 3 | 12,530 | 0.415 |
|  | Malawi | 1 | 3 | 12,530 | 0.232 |
| Rural | Vietnam | 3 | 4 | 2,486 | 0.485 |
|  | Malawi | 2 | 3 | 11,059 | 0.351 |
|  | Malawi | 1 | 3 | 11,059 | 0.282 |
| Urban | Ghana | 2 | 4 | 7,279 | 0.102 |
|  | Vietnam | 3 | 4 | 684 | 0.432 |
|  | Malawi | 2 | 3 | 1,471 | 0.454 |
|  | Malawi | 1 | 3 | 1,471 | 0.144 |

*Table S3. Summary of regression results. There are 3 levels of administrative units in Malawi, and 4 in each of Vietnam and Ghana, where enumeration area corresponds to the highest level of administrative unit. † In Malawi, the comparisons using CIESIN input data defined at the administrative levels 1 and 2 are relying on simulated population estimates, as described above.*

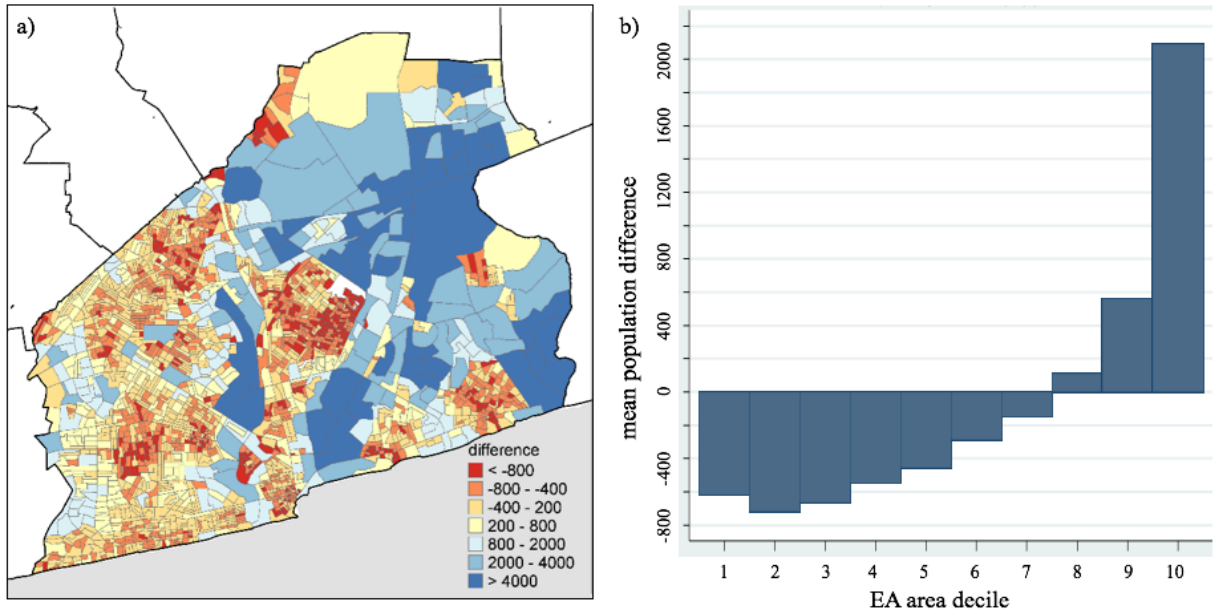

*Figure S2. Differences in population attribution (HRSL population estimate vs. EA population) versus EA size in the metropolitan area of Accra, Ghana. a) shows the difference by EA, note that the input census data for Ghana (the Parliamentary Constituency) is here indicated by the black line, i.e. only one unit of the Parliamentary Constituency is redistributed over 7279 Enumeration Areas. b) shows the same data, but grouped by EA area size, indicating a systematic under estimation in small EA areas in urban areas, likely due to the fact that our model does not take building height into account.*

### 5. Validation of building localization with household survey data

In this section, we assess the performance of the building identification at a granular level. To do so, we first overlay on the country's high-resolution population map the confidential GPS coordinates for the dwellings of the households interviewed by the Malawi Third Integrated Household Survey (IHS3)[8]. Subsequently, we compute the incidence of a Malawian household appearing in a pixel with no population according to the population map.

On the whole, there are minimal discrepancies between the high-resolution population and the household survey data. The incidence of finding Malawian households at a distance of more than 100m from the nearest populated pixel was only 1.7 percent at the national-level. In computing this statistic, we use a 100-meter tolerance in view of the (i) possible limitations in accuracy of

---

[8] The Third Integrated Household Survey (IHS3) was conducted by the Malawi National Statistical Office (NSO) from March 2010 to March 2011, with financial and technical support from the World Bank Living Standards Measurement Study – Integrated Surveys on Agriculture (LSMS-ISA) initiative. The IHS3 data are representative for key socioeconomic outcomes at the national-, urban/rural-, regional-, and district-levels. The sample includes 12,271 households, distributed across 768 EAs from 31 districts. The IHS3 anonymized unit-record data and documentation are publicly available on www.worldbank.org/lsms. The unit-record IHS3 data that are publically available include EA-level GPS coordinates that are computed as an average of household GPS coordinates in each EA and that are randomly off-set to preserve community and respondent confidentiality by 0-2 kilometers in urban areas and 2-5 kilometers in rural areas. The confidential georeferenced dwelling locations that we use for our analyses are not publically available. The research clearance was provided by the Malawi NSO, who, along with the World Bank Living Standards Measurement Study (LSMS) are the sole gatekeepers of the confidential data, which are not shared with anyone else inside or outside the World Bank, including the co-authors affiliated with Facebook and CIESIN.

GPS position, (ii) GPS measurements having been taken outside of the structures, and (iii) additional imprecision introduced by the gridding process itself (up to 1 gridcell of 30m). The incidence of a Malawian household appearing in a pixel with no population according to the high-resolution population map is 0.4 percent if one defines positive household identification as being within 500 meters of a populated pixel. To better understand the profiles of the "missed" IHS3 households, we reviewed each missed location manually on the Digital Globe imagery, and observed that these households were often 1) frequently occurring in clusters of three or more, and 2) more remote and with smaller structures compared to the rest of the sampled EA. The former may be linked to the mismatch between the dates of the imagery and the dates of the IHS3 fieldwork.

6. **Comparison with GUF and GHSL**

Two independent classifications of settlements are used for validation of the binary classification: the Global Urban Footprint (GUF) [7] and the Global Human Settlement Layer (GHSL) built-up grid [8]. Both comparison classifications use different data sources and methods to identify settlements. The GUF data represent settlements detected by synthetic aperture with an unsupervised classification scheme applied to backscattering amplitude and texture information from RADAR data—TerraSAR-X/TanDEM-X image products collected in 2011 and 2012[9] at a resolution of 0.4 arc-seconds, approximately 12 metres at the equator [7,9], The GHSL data represent settlements detected by a supervised classification method based on symbolic machine learning from optical imagery—Landsat satellite data collected in 2014 at a resolution of 38 metres [8]. These independent classifications were chosen because of the similar spatial resolution, the proximity in time to the building dataset and the differing data sources.

The three datasets are compared at the resolution of the building dataset —1 arc-second—to create all possible combinations of absence or presence of a settled area. Single-pixel areas of disagreement and areas of cloud in the source DigitalGlobe imagery were removed. The remaining areas where our classification disagrees with both other classifications represented as contiguous vectors are compared with the DigitalGlobe optical imagery to identify false positive and negative units. Country totals of agreement and disagreement as a proportion of country area are shown in Table S4**.**

---

[9] Data gaps for approximately 7% of the global data are filled with more recent data from 2013 and 2014 (DLR, 2016)




\begin{thebibliography}{10}
\bibitem{policy}
{By combining satellite imagery with census data, there is no association with
  any information that would lead to re-identification of individuals, and the
  granularity of the data is not enough for individual identification.
  Additionally, no Facebook user data has been used in this project.}
\end{thebibliography}
\end{document}